# Comparative Analysis of Probabilistic Models for Activity Recognition with an Instrumented Walker

**Farheen Omar, Mathieu Sinn, Jakub Truszkowski,
Pascal Poupart, James Tung and Allan Caine**[*]
David R. Cheriton School of Computer Science
University of Waterloo, Waterloo, Ontario, Canada
{f2omar,msinn,jmtruszk,ppoupart,j6tung}@uwaterloo.ca, adcaine@alumni.uwaterloo.ca

## Abstract

Rollating walkers are popular mobility aids used by older adults to improve balance control. There is a need to automatically recognize the activities performed by walker users to better understand activity patterns, mobility issues and the context in which falls are more likely to happen. We design and compare several techniques to recognize walker related activities. A comprehensive evaluation with control subjects and walker users from a retirement community is presented.

## 1 Introduction

Improving the quality of life of the ever increasing elderly population is one of the key concerns for health care provision. Limitations to independent mobility for these individuals have a significant impact on the quality of their life. Devices such as rollating walkers are often used to improve the independence and mobility of older adults. Our long-term goal is to improve the utility of these devices, by enabling them to perceive their environment and actively provide assistance to their users.

We are collaborating with a multidisciplinary group that is studying the usage of rollating walkers. We have access to a walker [9] that has been instrumented with various sensors and cameras to monitor the user. We are developing automated techniques to recognize the activities performed by users with respect to their walker (e.g., walking, standing, turning, etc.) based on the non-video sensors. This problem is significant for Kinesiologists who are studying the usage of walkers by elderly people. Currently they have to hand label the data by looking at video feeds of the user, which is not only time consuming but may not be accurate due to synchronization issues between various sensors. An automated activity recognition system would enable clinicians to gather statistics about the activity patterns of users, their level of mobility and the context in which falls are more likely to occur. This will also be useful for the development of smart walkers that can assist users with navigation and braking while taking into account their intended activity.

In this paper we describe a comparative analysis of activity recognition techniques based on hidden Markov models (HMMs) and conditional random fields (CRFs) trained by supervised and unsupervised learning for rollating walkers instrumented with various sensors. Our contributions are:

- the first fully automated system to automatically recognize activities performed by walker users;

- design and training (supervised and unsupervised) of probabilistic models (HMMs and CRFs) tailored to activity recognition with instrumented walkers;

- comprehensive analysis of these techniques with real data collected with control subjects and regular walker users living in a retirement community;

- comprehensive analysis of the ease/difficulty to recognize common walker user activities with existing algorithms.

The paper is organized as follows. Section 2 summarizes related work. Section 3 describes the walker, the experimental setup and our hypotheses regarding the ease/difficulty of recognizing common user activities. Section 4 describes the recognition models (HMMs and CRFs) and their training procedures (supervised and unsupervised). Sections 5 and 6 present the results of the experiments and analyze each approach. Finally Section 7 concludes and discusses future work.

---

## 2 Related Work

In [1], Alwan et al. describe a method that assesses basic walker-assisted gait characteristics, including heel strikes, toe-off events, as well as stride time, double support and right and left single support phases. These statistics are based on the measurement of weight transfer between the user and the walker by two load cells in the handles of the walker. A simple thresholding approach is used to detect peaks and valleys in the load measurements, which are assumed to be indicative of certain events in the gait cycle. This work focuses on low level gait statistics where as we are interested to recognize complex high level behaviours.

Hirata et al. [3] instrumented a walker with sensors and actuators. They recognize three user states: walking, stopped and emergency (including falling). These states are inferred based on the distance between the user and the walker (measured by a laser range finder) and the velocity of the walker. This work is limited to the three states mentioned above and would not be able to differentiate between activities that exhibit roughly the same velocity and distance measurements (e.g., walking, turning, going up a ramp).

A significant amount of work has been done on activity recognition in other contexts. In particular Liao et al. [6] use a Hierarchical Markov Model to learn and infer a user's daily movements through an urban community. The model uses multiple levels of abstraction in order to bridge the gap between raw GPS sensor measurements and high level information such as a user's destination and mode of transportation. They use Rao-Blackwellized particle filters for state estimation. In [5], they also recognize activities and places from GPS traces by Hierarchical CRFs. This work assumes that the high level goals and routines of the user as well as the map of the area are known and stable.

## 3 Experimental Setup

We have access to a walker developed by Tung et al. [9]. A picture of the walker is shown in Fig. 2. The walker is equipped with various sensors including a 3-D accelerometer in the seat, a load-cell in each leg and a wheel encoder, which measures the wheel's displacement. The sensor readings vary between 0 and $2^{16}-1$. The data is channeled via blue-tooth to a PDA for acquisition at 50 Hz. In addition to these sensors, there are two cameras on the walker. One is facing backwards and provides the video feed of the user's legs. The other is looking forwards and provides the video feed of the environment. The video frame rate is approximately 30 frames/second. In order to collect data for our models, we designed and conducted two exper-

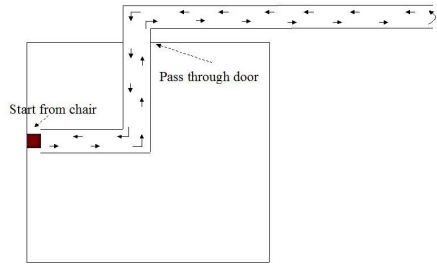

Figure 1: Course used for healthy young subjects

Table 1: Behaviours in Experiment 1

| Not Touching the Walker (NTW) |
| --- |
| Stop/Standing (ST) |
| Walking Forward (WF) |
| Turn Left (TL) |
| Turn Right (TR) |
| Walking Backwards (WB) |
| Transfers (Sit to Stand/Stand to Sit) (TRS) |

iments that are described below:

### 3.1 Experiment 1

17 healthy young subjects (age between 19 and 53) were asked to go through the course shown in Figure 1 twice with the walker. The behaviours exhibited by the participants are shown in Table 1.

### 3.2 Experiment 2

In a second experiment, we asked 8 regular walker users (age 84 to 97) to follow the course shown in Figure 3. This experiment was conducted at the Village of Winston Park (retirement community in Kitchener, Ontario) and the participants were residents of that facility. We also asked 12 adults (age 80 to 89) who do not live in a retirement community and are not regular walker users to follow the same course. The behaviours exhibited during this experiment include those of Tables 1 and 2.

### 3.3 Recognizing Behaviours

Our goal is to perform behaviour recognition based on the non-video sensors.[1] Since the accelerometers, load cells and wheel encoder only measure indirectly the activities of the person, it is not obvious a priori which activities can easily recognized. We formulated the following hypotheses.

---

[1] The use of video data is subject to future work.

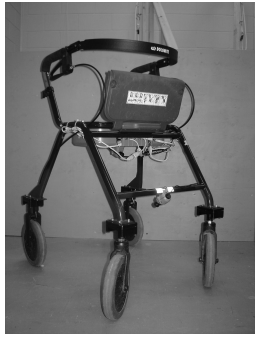

Figure 2: Smart Walker

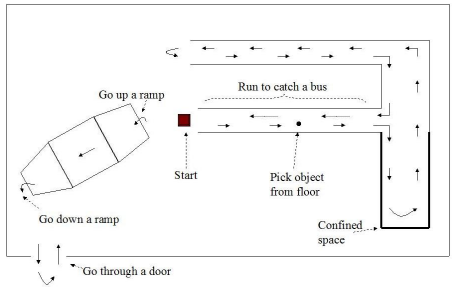

Figure 3: Course for older walker users

Not Touching Walker (NTW) should be easy to predict as there is no load on the walker and the walker is not moving. Standing (ST) should be differentiable from NTW based on load cell readings as load fluctuations are expected when the person touches the walker. Walking forward (WF) and walking backward (WB) should be differentiable from ST based on the wheel encoder measurements, which increase with forward movements and decrease with backward movements. Sitting on walker (SW) should also be easily distinguishable since the value of the load sensors is much higher than any other behaviour.

We expect turns to be more difficult to predict. We hypothesize that the speed of the person is lower when he/she is turning. We also expect a higher load on the side of the turn and some mild acceleration in the opposite direction of the turn. However this will likely vary with each person. Similarly, going up or down ramps and curbs are not expected to be easy to detect. We expect to see some fluctuations in the vertical acceleration when there is an immediate change of elevation and a sustained mild acceleration corresponding to gravity in the forward or backward direction depending on the inclination of ramps. We hypothesize that both these behaviours will be noticeable. However, they may be difficult to distinguish from WF in general. Furthermore, going up and down curbs may be particularly difficult to recognize due to the wide range of strategies used by people to lift or lower the walker.

Transfers (sit to stand or stand to sit) are also expected to be difficult to recognize due to a wide range of strategies. In theory, the load of the walker should decrease as the person goes from standing to sitting and increase for sit to stand. However, some people leave the walker to hold other supports such as the arms of a chair. Some people also engage the walker's brakes as they do a transfer while others move the walker. Another difficult behaviour is the reaching task. It involves behaviours such as opening a door or picking something from ground. The wheel encoder is not useful as people's habits of engaging the brakes during these behaviours are variable. In reaching tasks, however, the person usually keeps one hand on the walker while he/she uses the other hand to reach for the object. This can be captured by the load cells as there will be more weight on one side.

In the next section, we discuss various probabilistic models to recognize activities as accurately as possible despite the wide range of strategies for some behaviours.

## 4 Activity Recognition Models

We assume that the set of all possible behaviours is $\mathcal{B}$ whose cardinality is $m$. The behaviour at time $t$ is represented by the random variable $B_t$. The reading on sensor $k$ at time $t$ is given by the random variable $S_t^k$ where $k \in \{1, \ldots, n\}$. For notational convenience, we denote the sequence of behaviours from time $i$ to $j$ by $B_{i:j}$ and the observation sequence on sensor $k$ by $S_{i:j}^k$. The readings on all sensors observed between time $i$ and $j$ is denoted by $S_{i:j}^{1:n}$ and the actual observed sequence is denoted by $s_{i:j}^{1:n}$. The total length of a sequence is $T$. The actual behaviour at time $t$ is denoted by $\hat{b}_t$ and the predicted behaviour at time $t$ is $\tilde{b}_t$. Lower case letters denote the assignment of a value to a random variable.

### 4.1 Hidden Markov Model

We construct a hidden Markov model in which the behaviour is the hidden variable and the sensor readings are the observations (Fig. 4). The parameters include $\theta_{b',b} \equiv \Pr(B_t = b'|B_{t-1} = b)$ (probability that the behaviour at time $t$ is $b'$ given that the behaviour at time $t-1$ is $b$), $\phi_{s,k,b} \equiv \Pr(S_t^k = s|B_t = b)$ (probability that the value measured by the $k^{th}$ sensor at time $t$ is $s$ given that the behaviour is $b$) and $\pi_b \equiv \Pr(B_0 = b)$ (probability that the initial behaviour is $b$).

Table 2: Additional Behaviours in Experiment 2

| Going up Ramp (GUR) |
| Going down Ramp (GDR) |
| Sitting on Walker (SW) |
| Reaching Task (RT) |
| Going up Curb (GUC) |
| Going down Curb (GDC) |

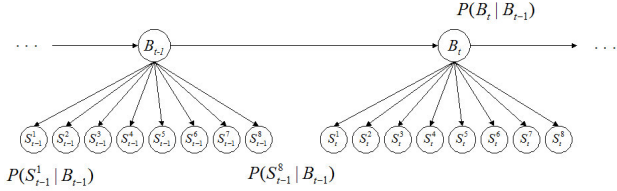

Figure 4: HMM for behaviour recognition

### 4.1.1 Maximum Likelihood (ML) Learning

For supervised learning, we manually label the data based on the video feed. We learn $\theta_{b',b}$ by counting the number of times behaviour $b$ is followed by behaviour $b'$ in the labeled data:

$$\theta_{b',b} = \frac{\sum_{t=1}^{T} \delta(B_t = b' \& B_{t-1} = b)}{\sum_{t=1}^{T} \delta(B_{t-1} = b)} \quad \forall b, b' \in \mathcal{B}$$

Here $\delta(x) = 1$ if $x$ is true and 0 otherwise. However, in some of our experiments, to avoid the bias introduced by using a fixed course, we use a simple transition model that reflects the fact that behaviours are $\tau$ times more likely to persist than to change.

$$\theta_{b',b} = \begin{cases} \tau/m + \tau - 1 & \text{if } b = b' \\ 1/m + \tau - 1 & \text{otherwise} \end{cases} \quad \forall b, b' \in \mathcal{B}$$

We can learn the prior $\pi_i$ from data using

$$\pi_b = \sum_{t=1}^{T} \delta(B_t = b)/T$$

Again, in some of our experiments, to avoid the bias introduced by using a fixed course, we consider all behaviours to be equally likely initially. Hence

$$\pi_b = 1/m \quad \forall b \in \mathcal{B}$$

We can model $\phi_{s,k,b}$ with a parametric density function such as a Gaussian. However, on close analysis of the data, we found that the distribution of $\phi_{s,i,j}$ does not follow a Gaussian. Therefore, we discretize the sensor readings by dividing the range into $D$ discrete intervals. We learn $\phi_{s,k,b}$ using

$$\phi_{s,k,b} = \frac{\sum_{t=1}^{T} \delta(B_t = b \& S_t^i = s)}{\sum_{t=1}^{T} \delta(B_t = b)} \quad \begin{array}{l} \forall b \in \mathcal{B}, \forall s \in \{1, \ldots, D\} \\ \forall k \in \{1, \ldots, n\} \end{array}$$

**ML with EM algorithm** For unsupervised learning, we use Expectation Maximization (EM) [2] to learn the parameters. The algorithm alternates between computing the expectations

$$E_0(b) = \Pr(B_0 = b | s_{1:t}^{1:n}, \pi, \theta, \phi),$$
$$E(b, b') = \sum_{t=1}^{T-1} \Pr(B_t = b, B_{t+1} = b' | s_{1:t}^{1:n}, \pi, \theta, \phi),$$
$$E_k(b, s) = \sum_{t=1}^{T} \Pr(B_t = b, S_t^k = s | s_{1:t}^{1:n}, \pi, \theta, \phi)$$

and updating the parameters

$$\pi_b = E_0(b),$$
$$\theta_{b',b} = E(b, b') / \sum_{b'} E(b, b'),$$
$$\phi_{s,k,b} = E_k(b, s) / \sum_s E(b, s).$$

**Prediction**: Note that since the sensor readings at any given time are conditionally independent given the behaviour at that time, therefore, $\Pr(s_t^{1:n} | B_t) = \prod_{i \in I} \phi_{s_t,i,b}$. We use maximum a posteriori filtering to infer the most likely behaviour given past observations: $\tilde{b}_t = \max_{b \in \mathcal{B}} \Pr(B_t = b | s_{1:t}^{1:n})$. This computation can be performed online as only past observations are used.

### 4.1.2 Bayesian Learning

Bayesian learning is an alternative to maximum likelihood learning. We start from a prior distribution and update it using Bayes rule to obtain the full posterior distribution over the variables of interest. By considering the full posterior over the parameters, we hope to avoid the over-fitting issues, often experienced by the EM algorithm. Unfortunately, in most cases the posterior does not have a tractable form, so we cannot sample from it directly. Consequently, we resort to sampling techniques based on simulating a Markov chain whose stationary distribution is the posterior distribution of interest.

A convenient choice of a prior distribution over the parameters is a product of Dirichlet distributions of the form $Dir(\theta_1, \ldots, \theta_k; c_1, \ldots, c_k) = \frac{\Gamma(\sum c_i)}{\prod \Gamma(c_i)} \prod \theta_1^{c_1} \theta_2^{c_2} \ldots \theta_k^{c_k}$ where $\sum_i \theta_i = 1$. Each Dirichlet is a prior for the corresponding multinomial distribution of transitions or emissions from some state. The values $c_i$ can be understood as counts indicating how many times a particular transition/emission has been observed. The Dirichlet distribution is a *conjugate prior* for the multinomial distribution. Therefore, if the hidden states are known, the posterior distribution of the parameters will also be a product of Dirichlets with the counts updated by the number of observed transitions or emissions. If the hidden states are not known, the posterior becomes a mixture of exponentially many products of Dirichlets, each product corresponding to one possible state path.

We use Gibbs sampling to estimate the posterior over hidden states and parameters. We repeatedly sample each variable from the conditional distribution given all the other variables. It can be shown that the resulting Markov chain converges to the joint distribution of all variables. Here, we use the *collapsed* Gibbs sampler [7] which samples the hidden states and integrates out the parameters to speed up the convergence.

Table 3: HMM results for Experiment 2 using center of pressure (COP) features. Behaviour persistence parameter: $\tau = 4000$. Window size is 25. Observations are accelerometer measurements, speed, frontal and sagittal center of pressure and total weight. Overall accuracy is 77%.

|     | NTW  | ST    | WF    | TL    | TR    | WB   | RT   | SW    | GUR  | GDR  | GUC  | GDC  | Accuracy % |
|-----|------|-------|-------|-------|-------|------|------|-------|------|------|------|------|------------|
| NTW | 821  | 7434  | 0     | 95    | 0     | 0    | 89   | 68    | 16   | 7    | 14   | 0    | 9.58       |
| ST  | 1393 | 35203 | 360   | 3103  | 550   | 2165 | 6831 | 595   | 355  | 61   | 771  | 251  | 68.17      |
| WF  | 239  | 2663  | 42297 | 6416  | 12486 | 763  | 3167 | 0     | 1283 | 657  | 618  | 828  | 59.23      |
| TL  | 41   | 756   | 1076  | 12165 | 516   | 1095 | 705  | 51    | 141  | 152  | 149  | 20   | 72.12      |
| TR  | 23   | 404   | 1743  | 784   | 10623 | 873  | 621  | 0     | 146  | 333  | 159  | 123  | 67.10      |
| WB  | 0    | 0     | 0     | 174   | 58    | 447  | 251  | 0     | 25   | 3    | 54   | 33   | 42.78      |
| RT  | 529  | 3837  | 516   | 1141  | 545   | 693  | 4227 | 25    | 221  | 88   | 263  | 202  | 34.40      |
| SW  | 15   | 151   | 33    | 171   | 61    | 0    | 106  | 70769 | 14   | 0    | 4    | 10   | 99.21      |
| GUR | 0    | 0     | 62    | 20    | 18    | 0    | 52   | 0     | 2382 | 153  | 154  | 157  | 79.45      |
| GDR | 0    | 20    | 55    | 176   | 124   | 26   | 29   | 0     | 24   | 2389 | 81   | 270  | 74.80      |
| GUC | 0    | 17    | 19    | 40    | 18    | 0    | 49   | 0     | 5    | 0    | 2939 | 0    | 95.21      |
| GDC | 0    | 20    | 27    | 15    | 171   | 0    | 67   | 0     | 32   | 35   | 22   | 1311 | 77.12      |

Since we use a Dirichlet prior, the posterior probability $\Pr(B_{1:T}, s_{1:T}^{1:n})$ can be computed analytically:

$$\Pr(B_{1:T}, s_{1:T}^{1:n}) \qquad (1)$$

$$= \int_{\pi\theta\phi} \Pr(B_{1:T}, s_{1:T}^{1:n} | \pi, \theta, \phi) \Pr(\pi, \theta, \phi) d\pi d\theta d\phi \qquad (2)$$

$$= c \frac{\prod_{B_1} \Gamma(\gamma_{B_1})}{\Gamma(\sum_{B_1} \gamma_{B_1})} \prod_b \frac{\prod_{b'} \Gamma(\alpha_{bb'})}{\Gamma(\sum_{b'} \alpha_{bb'})} \prod_{b,k} \frac{\prod_{s^k} \Gamma(\beta_{bs^k})}{\Gamma(\sum_{s^k} \beta_{bs^k})} \qquad (3)$$

where $\alpha$'s, $\beta$'s and $\gamma$'s are transition, emission and initial state counts. By taking $\Pr(B_t | B_{1:t-1}, B_{t+1:T}, s_{1:T}^{1:n}) = \Pr(B_t, B_{1:t-1}, B_{t+1:T}, s_{1:T}^{1:n}) / \sum_{B_t} \Pr(B_t, B_{1:t-1}, B_{t+1:T}, s_{1:T}^{1:n})$ and simplifying, we get:

$$\Pr(B_t | B_{1:t-1}, B_{t+1:T}, s_{1:T}^{1:n})$$
$$\propto \frac{\alpha_{B_{t-1}B_t}}{\sum_{b_t} \alpha_{B_{t-1}b_t}} \cdot \frac{\alpha_{B_t B_{t+1}}}{\sum_{b_{t+1}} \alpha_{B_t b_{t+1}}} \cdot \prod_{k=1}^n \frac{\beta_{B_t s_t^k}}{\sum_s \beta_{B_t s}}$$

By repeatedly sampling each hidden state according to the distribution above, we are guaranteed to converge to the posterior distribution $\Pr(B_{1:T} | s_{1:T}^{1:n})$.

Although we integrate out the parameters analytically, we can sample efficiently from the distribution over the parameters. For any assignment to $B_{1:T}$, the conditional distribution $\Pr(\pi, \theta, \phi | B_{1:T}, s_{1:T}^{1:n})$ is a product of Dirichlet distributions. Since the Gibbs sampler provides us with a way to sample from $\Pr(B_{1:T} | s_{1:T}^{1:n})$, sampling from $\Pr(\pi, \theta, \phi | s_{1:T}^{1:n})$ can be done efficiently.

Gibbs sampling can be used both for learning and prediction. Here, we only use it for learning to simplify the comparison to other methods.

### 4.2 Conditional Random Field

Conditional Random Fields (CRFs) are probabilistic models for segmenting and labeling sequential data [4]. We consider the special case of linear-chain CRFs. A CRF specifies the distribution of a sequence of labels ($B_{1:t}$) conditioned on a sequence of observations ($S_{1:t}^{1:n}$). In our experiments, the labels are the behaviour of the user, and the observations are the sensor measurements at each time. The probability of $B_{1:t} = b_{1:t}$ conditioned on $\mathbf{S_{1:t}^{1:n}} = \mathbf{s_{1:t}^{1:n}}$ is given by

$$P_\lambda(B_{1:t} = b_{1:t} \mid \mathbf{S_{1:t}^{1:n}} = \mathbf{s_{1:t}^{1:n}}) = \frac{1}{Z(\mathbf{s_{1:t}^{1:n}})} \times \left( \prod_{t=1}^T e^{\mu \cdot f(s_t^{1:n}, b_t)} \right) \times \left( \prod_{t=2}^T e^{\nu \cdot g(b_{t-1}, b_t)} \right) \qquad (4)$$

where $Z(\mathbf{s_{1:t}^{1:n}})$ is a normalizing constant, $f(s_t^{1:n}, b_t)$ is a state feature function (possibly vector-valued) with the corresponding weights $\mu$, and $g(b_{t-1}, b_t))$ is a transition feature function with the corresponding weights $\nu$. Intuitively, the feature functions allow to incorporate which observations and labels are likely to occur together. Usually, the feature functions are kept fixed and the weights are learned from training data. Given a sequence of labeled training data, the most common approach to find the model weights is minimizing the negative log-likelihood. Writing $\lambda$ for the stacked weights $(\mu, \nu)$, the objective function is given by

$$\mathcal{L}(\lambda) = -\sum_{t=1}^T \mu \cdot f\left(s_t^{1:n}, b_t\right) - \sum_{t=2}^T \nu \cdot g\left(b_{t-1}, b_t\right)$$
$$+ \log Z\left(\mathbf{s_{1:T}^{1:n}}\right) + \frac{\lambda^T \lambda}{2\sigma^2} \qquad (5)$$

where the last term on the right hand side is a shrinkage prior to penalize large weights. In our experiments we chose $\sigma^2 = 1$, However, we found that scaling $\sigma^2$ by factors up to 10 and $10^{-1}$, does not yield big differences in the accuracy of the resulting models. Since the objective function is convex, its unique minimum can be found using gradient-based search. The term $Z(\mathbf{s_{1:t}^{1:n}})$, which also depends on $\lambda$, can be efficiently evaluated using dynamic programming [8]. We use conjugate gradients to minimize the negative log-likelihood and stop training after 100 iterations.

Table 4: CRF results for Experiment 2 using center of pressure (COP) features. Window size is 25. Observations are accelerometer measurements, speed, frontal and sagittal center of pressure and total weight. Overall accuracy is 81%.

|     | NTW | ST | WF | TL | TR | WB | RT | SW | GUR | GDR | GUC | GDC | Accuracy % |
|-----|-----|-----|-----|-----|-----|-----|-----|-----|-----|-----|-----|-----|-----|
| NTW | 6674 | 1834 | 0 | 0 | 0 | 0 | 0 | 36 | 0 | 0 | 0 | 0 | 78 |
| ST | 1017 | 48677 | 1100 | 296 | 0 | 0 | 129 | 374 | 0 | 0 | 45 | 0 | 94 |
| WF | 0 | 2490 | 67738 | 744 | 270 | 0 | 25 | 112 | 0 | 0 | 38 | 0 | 95 |
| TL | 0 | 2081 | 5958 | 8528 | 97 | 0 | 72 | 117 | 0 | 12 | 2 | 0 | 51 |
| TR | 0 | 1517 | 10632 | 350 | 3129 | 0 | 62 | 34 | 0 | 69 | 39 | 0 | 20 |
| WB | 0 | 383 | 501 | 55 | 29 | 0 | 59 | 0 | 18 | 0 | 0 | 0 | 00 |
| RT | 387 | 7787 | 3071 | 327 | 24 | 0 | 634 | 46 | 11 | 0 | 0 | 0 | 5 |
| SW | 0 | 271 | 147 | 77 | 0 | 0 | 0 | 70839 | 0 | 0 | 0 | 0 | 99 |
| GUR | 0 | 173 | 1647 | 28 | 0 | 0 | 0 | 32 | 938 | 0 | 180 | 0 | 31 |
| GDR | 0 | 95 | 1753 | 0 | 575 | 0 | 0 | 0 | 0 | 720 | 41 | 10 | 23 |
| GUC | 0 | 436 | 375 | 0 | 89 | 0 | 122 | 0 | 0 | 0 | 2065 | 0 | 67 |
| GDC | 0 | 156 | 982 | 0 | 1 | 0 | 0 | 0 | 0 | 339 | 9 | 213 | 13 |

To predict behaviours, we consider the speed of the walker and the acceleration in x, y and z-direction. Instead of using the raw data of the load sensors, we consider the following measurements: the total load (which is just the sum of the loads on each wheel), the frontal plane center of pressure (which is the difference between the loads on the left and on the right side divided by the total load) and the sagittal plane center of pressure (the difference between the loads on the rear and front wheels divided by the total load).

Our state feature functions are based on thresholding: for each pair of behaviours $b, b'$ and each sensor $k$, we compare the actual observation to a fixed threshold value. If the value is exceeded, we add some model weight $\mu_{bb'k}^{(e)}$, otherwise, we add some weight $\mu_{bb'k}^{(n)}$. We choose the threshold values manually by a visual inspection of the data. If the labels $b, b'$ cannot be well discriminated by looking at the data from sensor $k$, the threshold is chosen as the average value from sensor $k$; later on, such "irrelevant" thresholds will be given very low weights in the training of the model.

We use a very simple transition model to avoid a bias towards certain transitions due to the design of the walker course. In particular, the transition feature function is given by $g(b_{t-1}, b_t) = \delta(b_{t-1} = b_t)$, hence the corresponding weight $\nu$ is a scalar.

Given the model parameters and an observation sequence, the predicted behaviour sequence $\tilde{b}_{1:T}$ maximizes the a-posteriori probability of $B_{1:T}$,

$$\tilde{b}_{1:T} = \arg\max_{b_{1:T}} P(B_{1:T} = b_{1:T} \mid \mathbf{S_{1:T}^{1:n}} = \mathbf{s_{1:T}^{1:n}})$$

where the maximization is over all label sequences of length $T$. Note that $\tilde{b}_{1:T}$ can be computed efficiently similar to the Viterbi algorithm for HMMs.

## 5 Results

We present results for the HMM and CRF for both experiments. We use leave one out cross validation for each round of training and prediction. Specifically, if we want to predict the behaviour sequence for a certain participant in Experiment 2, we learn the parameters from all other participants in Experiment 2. For Experiment 1, each person goes through the course twice. We included one instance of the course in the training data along with data from other participants while testing for the same person.

Since the range of sensor readings is too large (all integers from 0 to $2^{16} - 1$), we divide the range into 20 intervals and set their length in such a way that the same number of readings fall into each interval. Since the participants' weight varies, we also normalize the load cell readings as follows: $normalizedValue = (value - min)/(max - min)$.

Ground truth is established by hand labeling the data based on the video. This process is not perfect since the labeler can make mistakes while identifying behaviour transitions. For example, the labeler may interpret that a left turn started at time $t$, while the turn may actually start some time before $t$, but it only becomes evident in the video at time $t$. Therefore, in order to calculate our error, we introduce the concept of window size. If the window size is $x$, then for the behaviour at time $t$, if we find the same behaviour in the window between time $t - x$ and time $t + x$ in the predicted sequence, we count it as a correct prediction. We vary our window size from 0 to 50 in intervals of 5. If we make a correct prediction within a window width of 25, then we are only off by half a second. Since older people perform behaviours at a rate that is much slower than half a second, this may still be considered accurate.

Table 5: Experiment 1 percentage accuracy for each behaviour. NL means the normalized load values are used. COP implies that center of pressure feature is used instead of normalized load values. LOD means observation model is learnt from data and transition model uses $\tau$ for behaviour persistence. LOTD means the observation, transition and prior probabilities are learned from data. Window size is 25.

| Learning | Accuracy | | | | | | | |
|---|---|---|---|---|---|---|---|---|
| | NTW | ST | WF | TL | TR | WB | TRS | Total |
| Supervised HMM NL | 65 | 88 | 95 | 96 | 92 | 95 | 85 | 91 |
| Supervised HMM COP | 75 | 85 | 89 | 93 | 89 | 98 | 78 | 88 |
| Supervised HMM LOTD NL | 67 | 89 | 96 | 95 | 91 | 95 | 85 | 92 |
| Supervised CRF | 96 | 82 | 98 | 89 | 80 | 94 | 70 | 93 |
| Unsupervised HMM EM | 100 | 14 | 48 | 94 | 90 | 97 | 0 | 61 |
| Unsupervised HMM Gibbs | 100 | 72 | 95 | 66 | 72 | 44 | 17 | 83 |

Table 6: Experiment 2 percentage accuracy for each behaviour. NL means that normalized load values are used. COP implies that center of pressure feature is used instead of normalized load values. LOD means that the observation model is learnt from data and the transition model uses $\tau$ for behaviour persistence. LOTD means the observation, transition and prior probabilities have been learned from data. Window size is 25

| Learning | Accuracy | | | | | | | | | | | | |
|---|---|---|---|---|---|---|---|---|---|---|---|---|---|
| | NTW | ST | WF | TL | TR | WB | RT | SW | GUR | GDR | GUC | GDC | Total |
| Supervised HMM NL | 50 | 71 | 73 | 81 | 73 | 21 | 52 | 99 | 86 | 86 | 94 | 85 | 81 |
| Supervised HMM COP | 20 | 74 | 70 | 76 | 74 | 56 | 42 | 99 | 84 | 81 | 95 | 82 | 77 |
| Supervised HMM LOTD NL | 49 | 74 | 78 | 81 | 73 | 21 | 50 | 99 | 85 | 85 | 94 | 84 | 81 |
| Supervised CRF | 78 | 94 | 95 | 51 | 20 | 0 | 5 | 99 | 31 | 23 | 67 | 13 | 81 |
| Unsupervised HMM EM | 73 | 33 | 54 | 63 | 63 | 7 | 30 | 99 | 40 | 23 | 13 | 91 | 62 |
| Unsupervised HMM Gibbs | 100 | 42 | 77 | 69 | 81 | 0 | 62 | 91 | 57 | 52 | 46 | 2 | 69 |

Tables 3 and 4 show the results in the form of confusion matrices for Experiment 2 when performing supervised learning with an HMM and a CRF. In both cases, accelerometer measurements, speed, frontal and sagittal center of pressure (COP) and total weight are used as observations instead of the raw measurements (see Sect. 4.2 for more details). Each entry at row $i$ and column $j$ indicates how many times behaviour $i$ was confused as behaviour $j$, assuming a window of size 25.

Due to a lack of space, we did not include the confusion matrices for Experiment 1 and for the unsupervised learning algorithms, however Tables 5 and 6 summarize the recognition accuracy of all the algorithms for each experiment. We define accuracy as $\sum_{t=1}^{T} \delta\left(\hat{b}_t = \tilde{b}_t\right)/T$. Note that random predictions would yield an accuracy of $1/7 = 14\%$ in Experiment 1 and $1/13 = 7\%$ in Experiment 2.

In some situations, identifying transitions from some behaviour to another is what really matters. Hence, Fig. 5 shows the precision and recall of behaviour transitions (in addition to recognition accuracy) as a function of the window size. We calculate the number of actual transitions (i.e., $AT = \sum_{t=1}^{T} \delta\left(\hat{b}_t \neq \hat{b}_{t-1}\right)$), the number of predicted transitions (i.e., $PT = \sum_{t=1}^{|T|} \delta\left(\tilde{b}_t \neq \tilde{b}_{t-1}\right)$) and the number of correctly predicted transitions (i.e., $CPT = \sum_{t=1}^{T} \delta\left(\tilde{b}_t \neq \tilde{b}_{t-1} \& \hat{b}_t = \tilde{b}_t \& \hat{b}_{t-1} = \tilde{b}_{t-1}\right)$). Then $precision = CPT/AT$ and $recall = CPT/PT$.

## 6 Discussion

In this section, we analyze the results presented in the previous section.

**Experiment 1 vs. Experiment 2**: It is obvious from Tables 5 and 6 that the overall accuracy is much higher for Experiment 1. Difficult behaviours such as TR, TL and TRS are also predicted accurately. In fact, the accuracy for TRS is much higher than expected. One reason for the high accuracy may be that in Experiment 1, each person executes the course twice and one instance of the course execution is included in the training data while testing for the same person. Therefore, the training data may include information that is more specific to the particular person e.g., how people put load on the walker for different behaviours. Secondly, for Experiment 1, behaviours such as NTW, ST, WF and WB are easily distinguishable from each other. Confusion is only likely between WF, TL and TR and between ST and TRS. There is a larger number of behaviours in Experiment 2 that are difficult to distinguish from each other based on sensor information e.g. WF, TL, TR, GUR, GDR, GUC and GDC. Similarly it is difficult to distinguish between RT and ST.

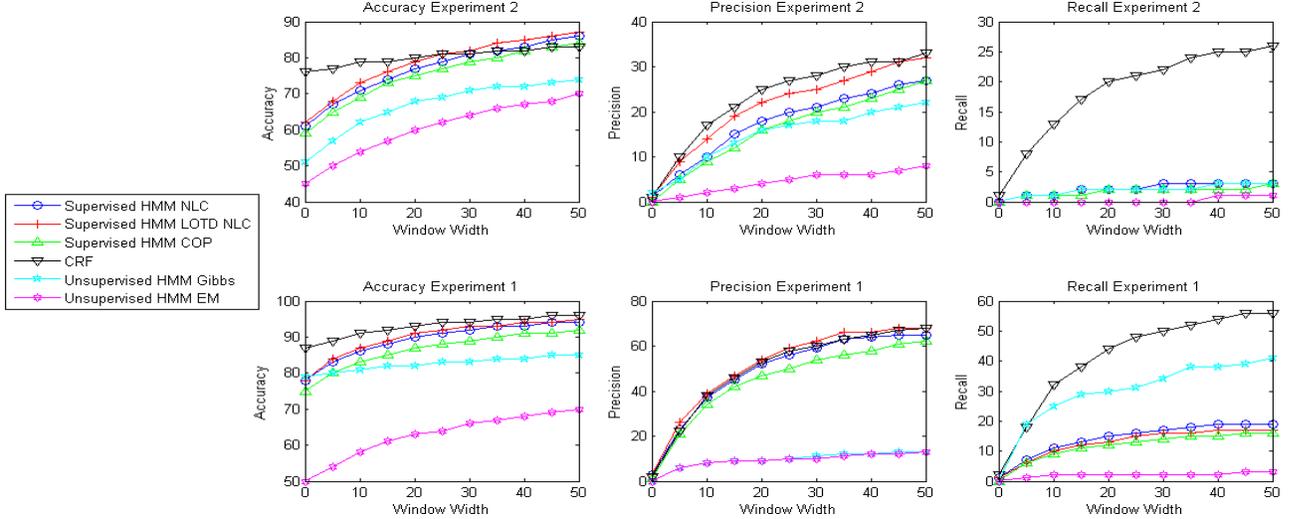

Figure 5: Accuracy, Precision and Recall for various algorithms in each experiment.

**Center of Pressure (COP) vs. Normalized Load Values**: In addition to the accelerometer values and the walker speed, we considered two alternative sets of features for prediction. One set includes the normalized load cell measurements while the other includes the frontal plane COP, the sagittal plane COP and the total load on the walker. It is evident from Tables 5 and 6 that both sets of features predict different behaviours well. When the COP features are used, we note that the prediction accuracy is lower for TR, TL, GUR, GDR, GUC and GDC and higher for the other behaviours.

**CRF vs. HMM**: Tables 5 and 6 show that the total accuracy of the CRF is much higher than that of the HMM. This is largely due to the fact that the CRF models $Pr(Behaviours|Observations)$ directly and optimizes the parameters of this distribution. On the other hand, the HMM models $Pr(Observations|Behaviours)$ as well as $Pr(Behaviours)$, and then uses Bayes rule to calculate $Pr(Behaviours|Observations)$. Therefore, the HMM model is more complex and the number of parameters that we have to learn is larger. Also, the HMM makes an explicit assumption about the conditional independence of sensor measurements over time. The CRF avoids this (problematic) assumption since it does not model any distribution over the observations. However, techniques for unsupervised learning are better established for HMMs than CRFs. This becomes an important advantage since we do not need to label data in unsupervised learning.

We were surprised to see that the HMM was able to predict certain behaviours better than the CRF. In general, the HMM seems to favor behaviours that occur infrequently. For example, in Table 3, we can see that the prediction accuracy of GUR and GUC is higher than that of WF. We expected that it would be difficult to accurately predict infrequent behaviours such as GUR and GUC. Note also that WF is often predicted as GUR and GUC. We suspect that this is due to the assumption of conditional independence between different sensors.

**Learning Transition Model from Data**: As discussed in Sect. 3, the experiments include a pre-defined walking course that biases the behaviour transitions. This is why we considered two scenarios when learning an HMM: i) fixed transition model where each behaviour is $\tau$ times more likely to persist than to transition to some other behaviour with uniform probability (denoted by LOD in Tables 5 and 6) and ii) learned transition model (denoted by LOTD). Naturally, the accuracy is higher when the transition model is learned from data. In future work, we plan to collect data with participants in their daily activities instead of a scripted walking course, which will allow us to learn realistic and personalized transition models.

**ML vs. Bayesian Learning**: As explained previously, manually labeling the data is a time consuming and error-prone. Unsupervised learning algorithms avoid this problem. However, they take longer to converge and the solutions are usually approximations to the optimal parameters. It is interesting to note from Table 5 that for Experiment 1, the Gibbs sampling accuracy for NTW and ST is actually higher than other algorithms. One important difficulty with unsupervised learning is state matching. On one hand, unsupervised learning algorithms may pick sub-behaviours of composite behaviours as a state. For example, the ramp transition can be broken up into getting on the ramp (corresponding to a blip in the vertical acceleration), and walking on the ramp. The algorithm may find a state that matches either one or both instead of the behaviours going up/down ramp. On the other hand, if two behaviours are similar, the algorithm might treat them as the same state. We observe that in Table 5, TRS is almost never predicted correctly. This may be due to the fact that TRS is very similar to ST and hence they are merged into one state.

For EM and Gibbs sampling, once a model is learned, we manually associate each latent state with the behaviour that is the most frequent. In general we used a number of latent states equal to the number of behaviours, except for Gibbs sampling in Experiment 1 (Table 5) where we used more latent states (11) than behaviours (7). As a result, the accuracy improved. In future work, we would like to investigate more thoroughly whether using a larger number of latent states generally improves the accuracy.

Since EM often gets stuck in local optima, we did 20 random restarts and showed the results of the model with the highest likelihood. In contrast, Gibbs sampling does not suffer from this problem. It is evident from Tables 5 and 6 as well as Fig. 5 that Gibbs sampling performs better than EM.

## 7 Conclusion and Future Work

This paper presented a novel and significant application of activity recognition in the context of instrumented walkers. We designed several algorithms based on HMMs and CRFs, and tested them with real users at the Village of Winston Park (retirement community in Kitchener, Ontario). A comprehensive analysis of the results showed that behaviours associated with walker usage tend to induce load, speed and acceleration patterns that are sufficient to distinguish them with reasonable accuracy. In the future, we would like to further improve the recognition accuracy, by using the video data and exploring various feature extraction techniques. We are also working on improving the battery life and memory capacity of the walker to collect data over long periods of time. In particular, this will allow us to loan the walker to users, record their daily usage and learn realistic and personalized behaviour transition models. Finally, we hope to turn this work into a clinical tool that can be used to assess the mobility patterns of walker users and the contexts in which they are more likely to fall.

### Acknowledgments

This work was supported by funds from CIHR, NSERC, the Ontario Ministry of Research and Innovation (Pascal Poupart's ERA) and the Canadian government (Mathieu Sinn's postdoctoral fellowship). We also thank the UW-Schlegel Research Institute for Aging and the Village of Winston Park as well as all the volunteers who participated in the experiments.